\documentclass[journal,twoside,web]{IEEEtran}
\usepackage{bm}
\usepackage{amsmath,amssymb,amsfonts}
\usepackage{algorithmic}
\usepackage{graphicx}
\usepackage{textcomp}
\usepackage{multirow}
\usepackage{url}
\usepackage{array}
\usepackage{cite}
\usepackage{verbatim}
\usepackage{color}

\title{LIAF-Net: Leaky Integrate and Analog Fire Network for Lightweight and Efficient  Spatiotemporal Information Processing}

\author{Zhenzhi~Wu, Hehui~Zhang, Yihan~Lin, Guoqi~Li, Meng~Wang, and~Ye~Tang
{\thanks{ The work was partially supported by National Key R\&D  program (2018YFE0200200), and Beijing Academy of Artificial Intelligence (BAAI), and a grant from the Institute for Guo Qiang of Tsinghua University, and key scientific technological innovation research project by Ministry of Education, and the open project of Zhejiang laboratory.}
\thanks{Z. Wu, H. Zhang, M. Wang and Y. Tang are with  Lynxi Technologies Co., Ltd., Beijing, P. R. China. (E-mail: zhenzhi.wu@lynxi.com).
H. Zhang, M. Wang and Y. Tang are also with Beijing University of Posts and Telecommunications, Beijing, China.
Y. Lin and G. Li are with Center for
 Brain-Inspired Computing Research, Beijing
  Innovation Center for Future Chip,
   Department of Precision Instrument, Tsinghua  University, P. R. China. (E-mail: liguoqi@mail.tsinghua.edu.cn).
   The Corresponding authors: Zhenzhi~Wu and Guoqi~Li.}%
}}

\begin{document}
	\maketitle

\markboth{Under Review: IEEE Transactions on Neural Networks and Learning Systems} {Wu
\MakeLowercase{\textit{et al.}}: LIAF-Net: Leaky Integrate and Analog Fire Network for  Lightweight and Efficient  Spatiotemporal Information  Processing}
\begin{abstract}
Spiking neural networks (SNNs) based on Leaky Integrate and Fire (LIF) model have been applied to energy-efficient temporal and spatiotemporal processing tasks. Thanks to the bio-plausible neuronal dynamics and simplicity, LIF-SNN benefits from event-driven processing, however, usually faces the embarrassment of reduced performance. This may because in LIF-SNN the neurons transmit information via spikes. To address this issue, in this work, we propose a Leaky Integrate and Analog Fire (LIAF) neuron model, so that analog values can be transmitted among neurons, and a deep network termed as LIAF-Net is built on it for efficient spatiotemporal processing. In the temporal domain, LIAF follows the traditional LIF dynamics to maintain its temporal processing capability. In the spatial domain, LIAF is able to integrate spatial information through convolutional integration or fully-connected integration. As a spatiotemporal layer, LIAF can also be used with traditional artificial neural network (ANN) layers jointly. Experiment results indicate that LIAF-Net achieves  comparable performance to Gated Recurrent Unit (GRU) and Long short-term memory (LSTM) on bAbI Question Answering (QA) tasks, and achieves state-of-the-art performance on spatiotemporal Dynamic Vision Sensor (DVS) datasets, including MNIST-DVS, CIFAR10-DVS and DVS128 Gesture, with much less number of synaptic weights and computational overhead compared with traditional networks built by LSTM, GRU, Convolutional LSTM (ConvLSTM) or 3D convolution (Conv3D). Compared with traditional LIF-SNN, LIAF-Net also shows dramatic accuracy gain on all these experiments. In conclusion, LIAF-Net provides a framework combining the advantages of both ANNs and SNNs for lightweight and efficient spatiotemporal information processing.
	\end{abstract}
	

	\section{Introduction}\label{sec:Introduction}
	\label{sec:introduction}
	\IEEEPARstart{I}{n} recent years, Leaky Integrate and Fire (LIF) neuron models have been investigated deeply  \cite{burkitt2006review} \cite{maass1997networks} \cite{mostafa2017supervised}. Benefited from its bio-plausible neuronal dynamics and simplicity, LIF has been employed to simulate spiking neural networks (SNNs) and applied in scenarios of energy efficient temporal and spatiotemporal processing \cite{mostafa2017supervised}  \cite{wongsuphasawat2012exploring} \cite{rullen2001rate} \cite{diehl2015unsupervised} \cite{vasilaki2009spike}. A large range of applications have been demonstrated  including image classification \cite{deng2009imagenet}, video processing \cite{hinton2012improving} \cite{hu2016dvs}, posture and gesture recognition \cite{zhao2014feedforward},\cite{lee2014real}, voice recognition \cite{schrauwen2008compact}, etc. Many  neuromorphic chips are specifically designed for implementing LIF model based SNNs (termed as LIF-SNN) with high energy efficiency    \cite{merolla2014million} \cite{pei2019towards} \cite{davies2018loihi}. In LIF-SNN, however, for each time step the information transmitted between two neurons can only be spike or nothing, and it is usually denoted as a binary value for simplicity. Since a binary value has less accuracy compared with an analog value, LIF-SNNs deployed on neuromorphic chips usually face the embarrassment of drop in performance.

Note that, although the communication scheme in LIF models with spike works efficiently in biological SNNs, it is not difficult to transmit an analog value instead of a binary spike for neuromorphic chips, just like the artificial neurons in deep Artificial Neural Networks (ANNs). On the other hand, the structure and neuronal dynamics in artificial neurons in ANNs are greatly simplified, resulting in a lack of ability in capturing the effects of membrane potential accumulating, action potential generation and membrane resetting, and these dynamic procedures are essential for temporal information processing.
Some artificial neural networks such as Long Short-Term Memory (LSTM) \cite{hochreiter1997long} and  Gated Recurrent Unit (GRU) \cite{cho2014learning} are skill at processing temporal information, which use recurrent connections and hidden state updates. These networks rely on gating instead of threshold firing. These gate signals require extra neuronal connections and weights than biology threshold-oriented mechanism.

Thus, the key motivation is as follows: Is there any solution available that enables lightweight  spatiotemporal processing to benefit from bio-plausible temporal processing mechanism, whereas keeps rich communication capability through analog values among neurons? In such case, how the spatiotemporal capability and efficiency can be attained?

To address these problems, two fundamental issues have to be solved. The first issue is that it is not always convenient to use spike trains as input and output for a neural network layer. On the one hand, most of the images and videos in current digital devices are represented as analog or multi-valued pixels. When process input data by LIF, a spike generator \cite{andreopoulos2015visual} or edge detector \cite{panda2018learning} has to be applied. Extra modules may be required for color processing \cite{falez2019unsupervised}. These modules introduce additional cost if SNN being applied in computer vision applications. On the other hand, in recent years, many newly introduced ANN layers are applied for building deep neural networks. Such as pooling \cite{krizhevsky2012imagenet}, residual connection \cite{he2016deep}, varieties of activations (ReLU \cite{krizhevsky2012imagenet}, PReLU \cite{he2015delving}), and normalization alternatives (Batch normalization \cite{ioffe2015batch}, Layer normalization \cite{ba2016layer}). It is  a pity that current SNNs rarely use these mechanisms for network construction, mostly due to the activation format difference (spike versus analog), therefore a coding format converter is necessary \cite{pei2019towards}. These converters make building a mixed network with LIF layer and ANN layers not efficient. Regarding this issue, including the redundancy of introducing sensor data converter and inconvenience  of using ANN layer, we are motivated to design a more friendly interface for LIF.

The second issue is how to avoid performance degradation when implementing simple and bio-plausible LIF-SNNs\cite{DBLP:journals/nn/DengWHLDLZLX20}. For example, the performance of LIF-SNN  is not as good as the ANN counterparts either through ANN-SNN conversion \cite{rueckauer2016theory} or training directly \cite{lee2020enabling}, mostly due to the spiking (binary activation) \cite{roy2019towards}. Although it can be compensated by a temporal coding scheme and using longer temporal time steps (more than 100 \cite{lee2020enabling}). To increase the accuracy, more time steps are required \cite{lee2019enabling} whereas more synaptic integration computation would be consumed. For instance, a time interval higher than ten is necessary for a counterbalanced performance on recognizing a CIFAR-10 image compared with pure ANN with the same network topology \cite{wu2019direct}, which makes the computational complexity one or several orders of magnitude higher when event-driven is not enabled. For event-driven case, it still not easy to reach the equivalent computation workload as ANN when the spike is not sparse enough (reach 1\% sparsity for 100 time steps, then SNN has equivalent dendritic workload to ANN).

Motivated by the above-mentioned issues, in this work, we try to overcome the shortcomings of traditional LIF and propose a refined LIF model, which enhances the information transmission capability, maintains the LIF temporal dynamics, enables the friendly integration with ANNs and vision sensors, and maintains the spatial performance with limited time intervals. We propose  a Leaky Integrate and Analog Fire (LIAF) neuron model and build a  deep neural network  based on LIAF, termed as LIAF-Net, for efficient spatiotemporal processing. To allow more information transmitted among neurons, LIAF-Net uses analog values to represent neural activations instead of the traditional binary values in LIF-SNN.
In the temporal domain,  LIAF  keeps traditional LIF dynamics and this enables its temporal processing capability.
In the spatial domain, since the LIAF data format is fully compatible with traditional  ANNs,  LIAF-Net can be built with LIAF and other ANN layers together easily.
Thus, deep learning designers can use training framework like Tensorflow to build such network, just considering LIAF as a spatiotemporal layer like 3D Convolution (Conv3D) or Convolutional LSTM (ConvLSTM).

For verifying the performance of LIAF-Nets, we have conducted several experiments on various temporal and spatiotemporal datasets and compared them with traditional networks. The experimental results show that LIAF-Net reaches comparable performance to the networks built by GRU, LSTM on the bAbI Question Answering (QA) tasks with much less storage and computational cost than GRU or LSTM network. LIAF-Nets also reach state-of-the-art performance on several spatiotemporal Dynamic Vision Sensor (DVS) datasets, including MNIST-DVS, CIFAR10-DVS  and DVS128 Gesture with much less number of synaptic weights and computational overhead compared to ConvLSTM network or Conv3D network.
We compared the network structure of LIAF and several recurrent models and reveals that neither input gate nor output gate is used in LIAF, and therefore huge power for storage and computation can be avoided. In conclusion, LIAF-Net provides a framework combining the advantages of both ANNs and SNNs for lightweight and efficient spatiotemporal information processing.

The paper is organized as follows. In Section \ref{sec:Model Description}, the model description of the proposed LIAF is introduced. The relationships of LIAF with Perceptron, Convolution, RNN model and LIF are discussed. Also, the way for integrating LIAF layer with other ANN layers is proposed. In Section \ref{sec:Training LIAF}, the training algorithm is discussed. In Section \ref{sec:LIAF as a lightweight spatiotemporal model} and Section \ref{sec:Efficiency Evaluation} LIAF is compared with traditional models to show its lightweight characteristic. In Section \ref{sec:Performance Evaluation}, the performance of LIAF-Net is evaluated and compared with other traditional networks, which reveals the computational efficiency and outstanding spatiotemporal processing capability of LIAF-Net. In Section \ref{sec:Discussions}, the bio-plausibility of LIAF and analog valued neural models are discussed. Finally, Section \ref{sec:Conclusion} concludes the paper.

\section{Model Description}\label{sec:Model Description}
\subsection{Evolution from ANN and SNN models}
\begin{figure}[!t]
{\includegraphics[width=\columnwidth]{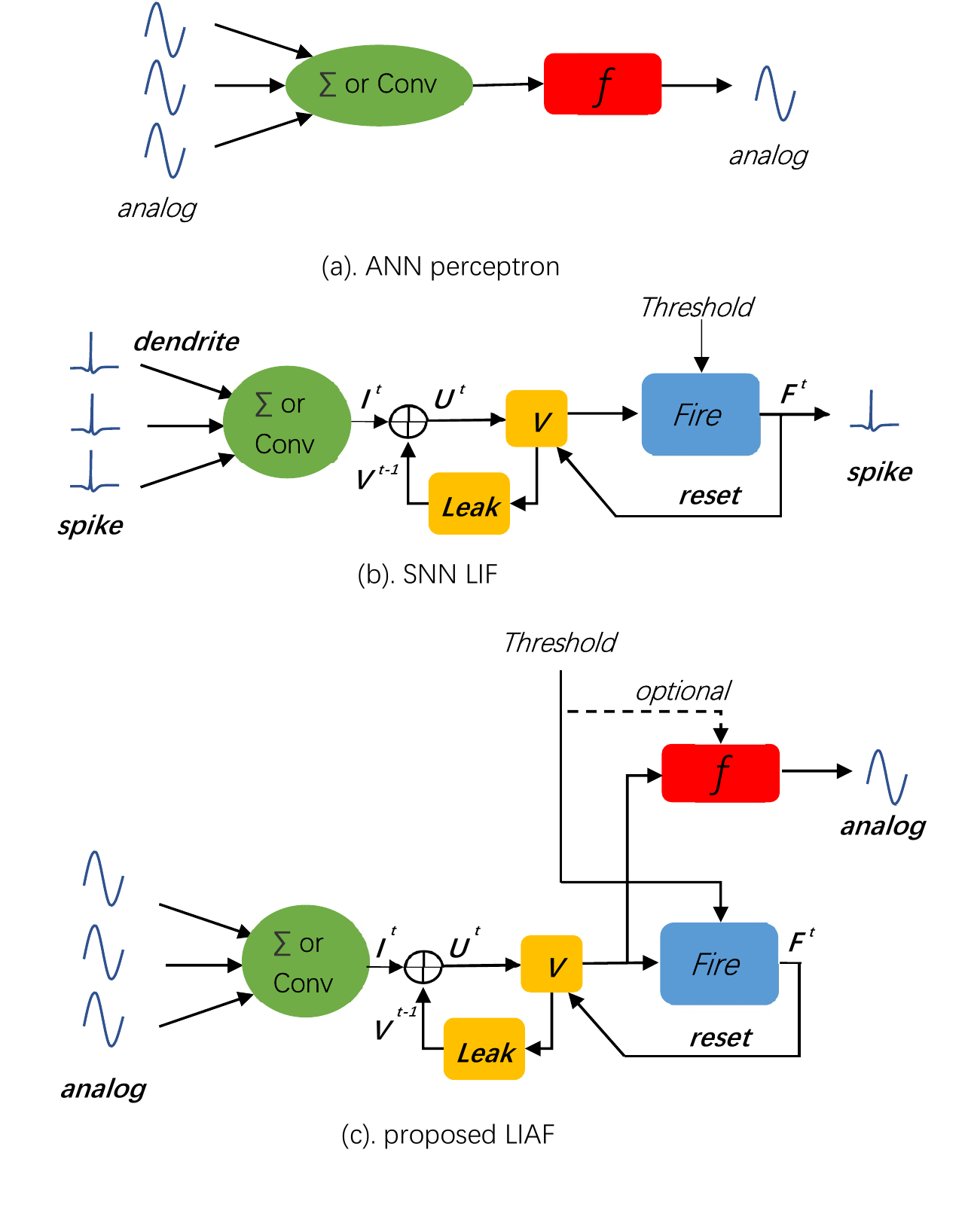}}
\caption{The comparison of traditional perceptron, LIF and proposed LIAF neuron models. The proposed LIAF keeps analog input and analog output like perceptron neuron, whereas maintains temporal dynamics similar to LIF neuron.}
\label{fig1}
\end{figure}

The proposed LIAF model is quiet similar to the LIF model. It has a similar dendritic integration process, temporal dynamics including threshold compare and fire, and membrane potential reset. However, it receives analog input and outputs analog activations, illustrated in Fig.\ref{fig1}. Different from LIF, the fire signal in LIAF is only used for resetting the membrane potential, therefore the membrane potential updates like the LIF model, whereas the output activations are calculated from the membrane potential through an activation function.

\subsection{Mathematical  description of the LIAF model}\label{sec:Formal Description}
The original LIF model is described in a differential function \cite{ferre2018unsupervised} \cite{roy2019towards} to reveal the neuronal dynamic, following equation
\begin{equation} \tau \frac{dV(t)}{dt}=-(V(t)-V_{reset})+\sum_{i=1}^n W_{i}\cdot X_{i}(t).\label{eq}\end{equation}
where $\tau$ is the timing factor of the neuron, $V_{reset}$ is the reset potential. $X_{i}(t)$ is the input signal (spike or none) from the ith neuron connecting to the current neuron through a synapse with strength $W_{i}$.
When $V(t)$ reaches a certain threshold $V_{th}$, a spike is emitted, and the $V(t)$ is reset to its initial value $V_{reset}$.
We use an iterative representation in discrete-time \cite{wu2018spatio} \cite{neftci2019surrogate} for easily inference and training. We present LIF and LIAF neuron model together for comparing as follows.

a). Synaptic Integration

\begin{equation}
\bm{I}^{t}=\begin{cases}
\bm{X}^{t}\cdot \bm{W}, & \mbox{for dense;} \\
Conv (\bm{X}^{t},\bm{W}), & \mbox{for convolution;} \\
\bm{X}^{t}, & \mbox{for integration-free.}
\end{cases}
\end{equation}

where $\bm{X}^{t}$ stands for the activations from the pre-synaptic neuron, $\bm{W}$ refers to the synaptic weights. The synaptic integration can be exist in fully connected format or convolutional format.

b). Accumulates with previous membrane potential
\begin{equation}\bm{U}^{t}_{m}=\bm{I}^{t}+\bm{V}^{t-1}_{m}\end{equation}

where $\bm{V_m}^{t-1}$ and $\bm{U_m}^{t}$ refer to the previous and current membrane potential respectively.

c). Compare with the threshold and fire
\begin{equation}\bm{F}^{t}=\bm{U}^{t}_{m}\ge \bm{V}_{th}\end{equation}

where $\bm{F}^{t}$ is the fire signal. For each $F_j^{t}$ in $\bm{F}^{t}$, $F_j^{t} = 1$ indicates a firing event, otherwise $F_j^{t}=0$.

d). Reset the membrane potential when fired
\begin{equation}\bm{R}^{t}_{m}=\bm{F}^{t}\cdot \bm{V}_{reset}+(\bm{1}-\bm{F}^{t})\cdot \bm{U}^{t}_{m}\end{equation}

e). Perform leakage
\begin{equation}\bm{V}^{t}_{m}=\bm{\alpha} \cdot \bm{R}^{t}_{m}+\bm{\beta} \end{equation}

where $\bm{\alpha}$ and $\bm{\beta}$ represent the multiplicative decay and additive decay respectively.

f). Output
\begin{equation}\bm{Y}^{t}=\begin{cases}
\bm{F}^{t}, & \mbox{for LIF;} \\
f (\bm{U}^{t}_{m},\bm{V}_{th}), & \mbox{for LIAF.}
\end{cases}\end{equation}

$ f (x,V_{th})$ is the analog activation function. It can be threshold related (TR mode) or not (NTR mode), which follows
\begin{equation}f (x,V_{th})=\begin{cases}
Act (x-V_{th}), & \mbox{for TR mode;} \\
Act (x), & \mbox{for NTR mode.}
\end{cases}\end{equation}

In the following illustrations, we term LIAF/LIF with dense integration, convolutional integration, and integration-free as DenseLIAF/DenseLIF, ConvLIAF/ConvLIF and DirectLIAF/DirectLIF respectively. The 'Conv' mentioned here is the 2D Convolution (Conv2D). In addition, units in $\bm{V}_{th}$, $\bm{V}_{reset}$, $\bm{\alpha}$ and $\bm{\beta}$ may vary for each convolutional channel (the neurons within a channel share a unique value), or vary for each neuron, or the same for all neurons, are termed as  Channel-Sharing mode, Non-Sharing mode and All-Sharing mode, respectively. We avoid using Non-Sharing mode in ConvLIAF/ConvLIF in our experiments for saving parameter storage. We also would like to note that the term 'model' is equivalent to a 'layer' using the model in a network.

\subsection{Model degradation}\label{sec:Model Degradation}
In this part, we compare LIAF to LIF, perceptron, convolution, and traditional recurrent neural network (RNN) models.

\textbf{(1). Relationship to LIF:}
If $Act(x)$ is a Heaviside step function ($Act (x) = V\ge V_{th}$), then LIAF will degrade to LIF.

\textbf{(2). Relationship to perceptron or convolution:}
If we set \textbf{$ \bm{\alpha}=\bm{0}$} and \textbf{$ \bm{\beta}=\bm{0} $}, then all the temporal information is missing and LIAF will degrade to perceptron or Conv2D.

\textbf{(3). Relationship to RNN model:}
We rewrite the LIAF as
\begin{equation}
\begin{split}
\bm{U}^{l,t}_{m}=\bm{I}^{t}+\bm{\alpha} \cdot S (\bm{U}^{l,t-1}_{m}-\bm{V}_{th}) \cdot \bm{V}_{reset}+ \\
\bm{\alpha} \cdot [1-S (\bm{U}^{l,t-1}_{m}-\bm{V}_{th})] \cdot \bm{U}^{l,t-1}_{m} + \bm{\beta}
\end{split}
\end{equation}
which can be viewed as a high order recurrent neural network because of the term $ S (\bm{U}^{l,t-1}_{m}-\bm{V}_{th}) \cdot \bm{U}^{l,t-1}_{m}$. Therefore it consists of more non-linear features compared with traditional RNNs. The comparison with traditional RNN models (LSTM and GRU) is discussed in Section \ref{sec:LIAF as a lightweight spatiotemporal model}.
In conclusion, the proposed model has the equivalent expressional capability to both ANN perceptron/convolution model and the LIF model.

\subsection{Easily integration with ANN layers}\label{sec:Easily integration with ANN layers}
\begin{figure}[!t]
\centerline{\includegraphics[width=\columnwidth]{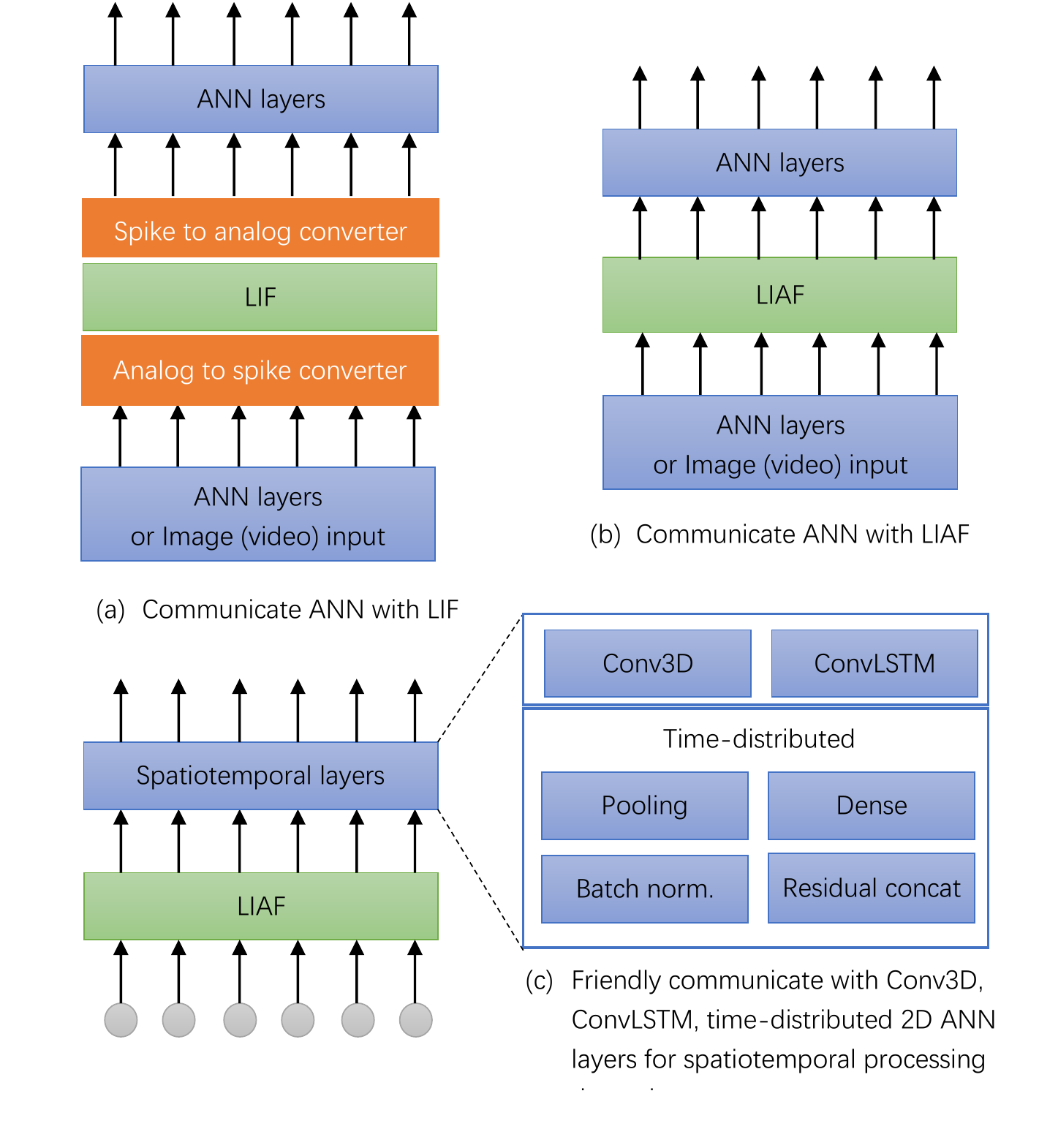}}
\caption{LIAF layer integrated with ANN layers. LIF usually requires a spike train other than a single spike to compensate for the accuracy loss. Therefore, when communicates with ANN layers with analog values, a converter between spike train and the analog value is required. Otherwise, only binary input is acceptable and the output accuracy is limited. For LIAF, no spike/analog conversion is required, and a mixed network construction approach with LIAF and traditional ANN layers is supported. (Note that the LIAF layer here only includes the processing steps mentioned in Section \ref{sec:Formal Description}, excluding other layers like pooling, batch normalization.) }
\label{fig2}
\end{figure}

Compared to LIF, a LIAF layer can be more easily integrated with traditional ANN layers easily. As depicted in Fig.\ref{fig2}. For video or frame sequence processing, the inter-layer interface of ConvLIAF/Conv3D/ConvLSTM is a 5D tensor in size $(B, T, H, W, L)$, representing batch size, time steps, height, width, and feature maps respectively. The interface of LIAF/LSTM/GRU is a 3D tensor in size $(B, T, L)$, where $L$ is the number of neurons (hidden size). Therefore LIAF and other ANN layers share the same data format and interface format, which makes their integration easily, just considering LIAF as a spatiotemporal layer similar to Conv3D or ConvLSTM. On the contrary, LIF requires a spike train to represent a multi-valued pixel or an analog activation in a \textbf{single time step}, otherwise only binary value is allowed. Therefore, when accepting analog values from ANN layers or from image/video sensor, a converter from analog value to spike train is necessary. In addition, the spike train fired by LIF may need to be converted to an analog value to communicate with the subsequent ANN layer.

\section{Training the LIAF-Net}\label{sec:Training LIAF}
\begin{figure*}[!t]
\centerline{\includegraphics[width=400pt]{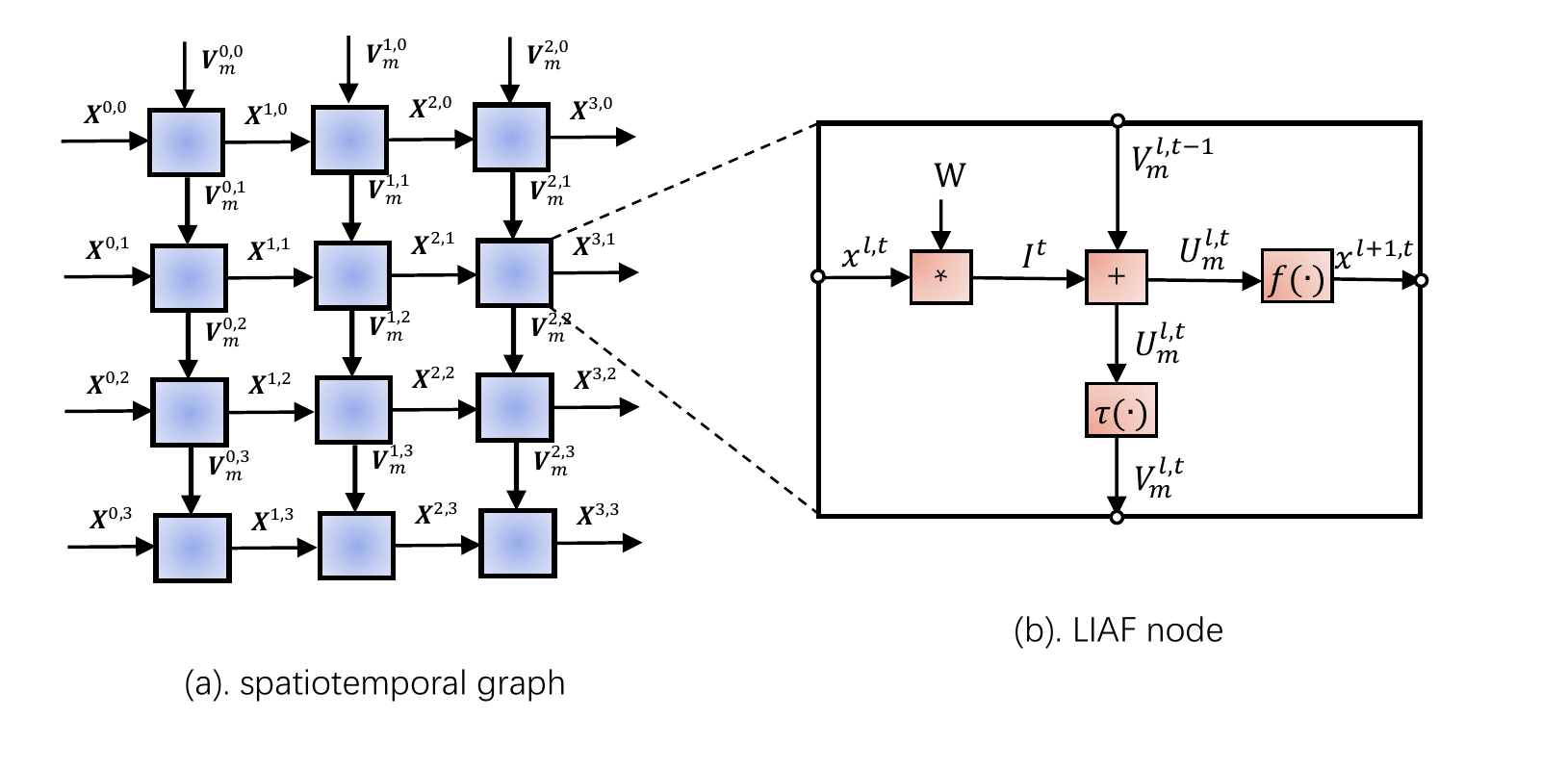}}
\caption{The spatiotemporal structure of LIAF-Net. LIAF-Net shares the same node-level 2D connections with RNN depicted in (a), whereas differs inside a node shown in (b). When training with Back Propagation Through Time (BPTT), LIAF follows similar back propagation chain as RNN.}
\label{fig3}
\end{figure*}

The LIAF-Net can be trained with Back Propagation Through Time (BPTT), which has been widely applied for the training of temporal RNNs. Since training an RNN and LIF-SNN with BPTT are well discussed in many literatures \cite{pineda1987generalization} \cite{neftci2019surrogate} \cite{lee2016training} \cite{hong2019training}, and implemented in mainstream deep learning frameworks, here we just discuss the difference between training RNN and LIAF-Net using BPTT. Both RNN and LIAF-Net are recurrent networks, therefore we unfold the time domain and form a 2D grid, shown in Fig.\ref{fig3} (a). Each of the node represents a Node Function (NF) which is defined as
\begin{equation}(\bm{V}^{l,t}_{m},\bm{X}^{l+1,t})=NF (\bm{V}^{l,t-1}_{m},\bm{X}^{l,t})\end{equation}
where $l$ is the layer index, and $t$ is the current time step, $\bm{V}^{l,t}_{m}$ represents the membrane potential of neurons (hidden state), and $\bm{X}^{l,t}$ denotes the activations.

Since recurrent networks follow the same graph as Fig.\ref{fig3} (a). They are trained with the same back propagation rule above NF level. The difference is inside the node. Following the definition in Section \ref{sec:Formal Description}, we can detail the LIAF node function in Fig. \ref{fig3} (b). For LIAF, the relationship between $\bm{U}^{l,t}_{m}$ and $\bm{V}^{l,t}_{m}$ can be described as a $\tau (\cdot)$ function
\begin{equation}
\begin{split}
\tau (\bm{U}^{l,t}_{m})=\bm{\alpha} \cdot S (\bm{U}^{l,t}_{m}-\bm{V}_{th})\cdot \bm{V}_{reset} \\
+ \bm{\alpha} \cdot [1- S (\bm{U}^{l,t}_{m}-\bm{V}_{th})] \cdot \bm{U}^{l,t}_{m} +\bm{\beta}
\end{split}
\end{equation}
where $S (x)$ is the Heaviside step function. Following the BPTT algorithm, the derivative chain is applied, when performing the partial derivative of $\bm{V}^{l,t}_{m}$ over $\bm{U}^{l,t}_{m}$, we have
\begin{equation}
\begin{split}
\dfrac{\partial \tau (\bm{U}^{l,t}_{m})}{\bm{U}^{l,t}_{m}}=\bm{\alpha} \cdot \delta (\bm{U}^{l,t}_{m}-\bm{V}_{th}) \cdot (\bm{V}_{reset}-\bm{U}^{l,t}_{m}) \\
+ \bm{\alpha} \cdot (1-S(\bm{U}^{l,t}_{m}-\bm{V}_{th}))
\end{split}
\end{equation}
where in $\delta (x)$ is the Dirac Delta function, which is not friendly with the backpropagation, therefore we introduce a rectangle window approximation for the $\delta (x)$ function, defined as
\begin{equation}\bar{\delta} (x)=\begin{cases}
1, & \mbox{when } |x|< \mu \\
0, & \mbox{otherwise}
\end{cases}\end{equation}
where $\mu$ is a small positive number.
The partial derivatives of other operations in the LIAF-Net are easy to obtain and can be easily handled by the deep learning frameworks automatically.

\section{LIAF as a lightweight spatiotemporal model}\label{sec:LIAF as a lightweight spatiotemporal model}
In this section, the proposed LIAF is compared with several temporal/spatiotemporal models including RNN, GRU, LSTM, ConvLSTM, Conv3D, which shows the reason why LIAF can be viewed as a lightweight spatiotemporal model.
The formal representation of these models are listed in reference \cite{hochreiter1997long} \cite{cho2014learning} \cite{ wiki:RNN}, and depicted in Fig.\ref{fig4}. Formal definitions are provided in Appendix A. Here the hidden state (cells) $\bm c$ in each sub figure in Fig.\ref{fig4} is a vector with size $L$ or a tensor with size $H\cdot W \cdot L$.
FC refers to a fully connected operator, named $\bm y=\sigma(\bm W\bm x+\bm b)$, where the activation function $\sigma$ and bias $\bm b$ are optional. gFC is used for gate calculation, defined as $\bm g=\sigma([\bm W_x, \bm W_h]\cdot[\bm x, \bm h]^T + \bm b)$, which indicates that all the cells are participating the calculation of a gate $\bm g$.

The traditional RNN model (shown in Fig.\ref{fig4} (a)) contains a recurrent FC for hidden state information exchange, which is not applied in LIAF/LIF. LSTM and GRU shown Fig.\ref{fig4} (b) (c) are widely applied recurrent models. These models consist of gating units which are controlled by the input of current step $\bm{X}_{t}$ and the output of previous time step $\bm{h}_{t-1}$. In LSTM, the gating functions are used for input, output and hidden state update which requires introducing three gFCs. In GRU the input and hidden state update share one gate calculated by a gFC.

For DenseLIAF/DenseLIF, the fire, reset, and leak can be viewed as `gates', which control how much previous state information can be kept in the current state. Different from LSTM and GRU, the gate is only controlled by the cell neuron itself, and only hidden state information is considered. Neither the input gate nor the output gate is used in LIAF, therefore no gFC is used for the overall network. Since gFC contains a large number of weights, it  requires huge computational power for storage and calculation (multiplication-addition operation). Reducing gFC will lead to the reduction of implementation cost to a large extend.

For spatiotemporal models, such as ConvLSTM, the gFCs are changed to convolutional operations, denoted as $gConv(\bm x,\bm h)=\sigma(Conv2D(\bm W_x, \bm x)+Conv2D(\bm W_h, \bm h) +\bm b)$. Similar saving from ConvLSTM to ConvLIAF on parameter and computation are still sustainable due to the saving of these convolutions. In addition, for ConvLSTM, an additional convolution for input $h_{t-1}$ is required which consumes more computation and storage. For Conv3D, the convolution operations are performed in a 3D block, which requires several times of mulplication-addition calculation compared to LIAF which only requires Conv2D.

In conclusion, LIAF can be viewed as a lightweight spatiotemporal model. In addition, for a network, LIAF can be used together with these classical models to reach a balance of performance and complexity.

\section{Efficiency Evaluation}\label{sec:Efficiency Evaluation}
We summarized the computational overhead and number of parameters of LIF, LIAF, RNN, GRU \cite{chung2014empirical} , and LSTM \cite{hochreiter1997long}, according to equations (2)-(8), and equations defined in Appendix A. The result is shown in Table \ref{evtab1}. The calculation is based on a single time step since there is no difference among all time steps. The batch size is set to one. We denote the size of the input vector and cell vector as $K$ and $L$. Note that the activation functions can be realized by look-up tables. The computational overhead of look-up table, selection and comparison operation is much lower than multiplication and addition, so we do not list them in Table \ref{evtab1}. It can be concluded that LIAF consumes the same addition and weights to LIF, while due to the presence of analog signals, the multiplications of LIAF are slightly higher than that of LIF. There are two reasons as follows. Firstly, since the data received by LIF is binary, the matrix multiplication of equation (2) can be replaced by a selection operation. Secondly, for LIF, the multiplication of reset membrane potential in equation (5) can also be reduced to a selection operation. The event-driven property of LIF is not counted in since LIAF can also benefit from similar event-driven when using appropriate activation function (such as ReLU with a bias), and the activation sparsity is highly data-dependent. Importantly, both LIAF and LIF consume far less computational overhead and the number of parameters than GRU and LSTM.
	
In Table \ref{evtab2}, we calculated the computational overhead and the number of parameters of spatiotemporal layers. We fix the batch size to one. The hidden state and output have the same tensor size $(T, H, W, L)$, where $T$ represents the temporal dimension, $H$ and $W$ denote the height and width of the feature map respectively, and $L$ represents output channel size. $(I, J)$ denotes the convolution kernel size of Conv2D, ConvLIF, ConvLIAF, and ConvLSTM, $(U, I, J)$ denotes the convolution kernel size of Conv3D, and $K$ denotes input channels size. To facilitate the comparison of formulas, in Table \ref{evtab2}, we use $R$ and $Q$ to represent \begin{math} T \cdot H \cdot W \cdot L \end{math} and \begin{math} I \cdot J \cdot K \end{math}. As shown in Table \ref{evtab2}, the computational overhead of ConvLIAF is not significantly different from that of Conv2D (time-distributed). ConvLIAF shares the same additions and parameters with ConvLIF. Compared with Conv3D and ConvLSTM, ConvLIAF can save a mass of calculation quantity and storage overhead.

Note that in our following experiments, we may split ConvLIAF into Conv2D and DirectLIAF, and insert additional layers (such as normalization) between them. In such case the overhead of ConvLIAF is used to evaluate the combination of Conv2D layer and DirectLIAF layer. Similarly, metrics for ConvLIF are used for evaluating Conv2D + DirectLIF.

In short, we concluded that LIAF has advantages in terms of computational complexity and storage capacity. We will provide numerical results in the subsequent experiments to more intuitively reflect the efficiency advantages of LIAF.
	
In another aspect, the communication overhead is an important factor for building energy efficient neuromorphic implementations. In these devices, the activations are transmitted via Network on Chip using packets. Take TrueNorth \cite{7229264} as a reference implementation, wherein each spike is transmitted by a 32-bit packet. If an 8-bit analog value is appended, then the additional bandwidth is 20\%. A similar strategy is adopted in Tianjic chip \cite{pei2019towards}, wherein each packet can transmit either 8-bit analog value or spike. For LSTM, although the size of output activations is the same, it requires much more communication or memory access resources on internal components, which may lead to much higher overhead when mapping to the neuromorphic devices.

	\begin{table}
		\caption{Formulas for calculating the computational complexity and the weights of different temporal layers. }
		\setlength{\tabcolsep}{3pt}
		\begin{tabular}{|p{40pt}|p{60pt}|p{60pt}|p{60pt}|}
			\hline
			\textbf{layer}&
			\textbf{MULs}&
			\textbf{ADDs}&
			\textbf{Weights}
			\\
			\hline
			DenseLIAF&
			\begin{math} L \cdot (K + 1) \end{math}&
			\begin{math} L \cdot (K + 2) \end{math}&
			\begin{math} L \cdot (K + 1) \end{math}  \\
			\hline
			DenseLIF&
			\begin{math} L \end{math}&
			\begin{math} L \cdot (K + 2) \end{math}&
			\begin{math} L \cdot (K + 1) \end{math} \\
			\hline
			RNN &
			\begin{math} L \cdot (L + K) \end{math}&
			\begin{math} L \cdot (L + K) \end{math}&
			\begin{math} L \cdot (L + K + 1) \end{math}  \\
			\hline
			GRU&
			\begin{math} L \cdot 3 \cdot (L + K + 1) \end{math}&
			\begin{math} L \cdot (3 \cdot (L + K) + 1)\end{math}&
			\begin{math} L \cdot 3 \cdot (L + K + 1) \end{math} \\
			\hline
			LSTM&
			\begin{math} L \cdot (4 \cdot (L + K) + 3)\end{math}&
			\begin{math} L \cdot (4 \cdot (L + K) + 1)\end{math}&
			\begin{math} L \cdot 4 \cdot (L + K + 1)\end{math}  \\
			\hline

		\multicolumn{4}{p{250pt}}{Note: MULs, ADDs, and Weights refer to the number of multiplication operations, addition operations and weights (including bias), respectively.}
		\end{tabular}
		\label{evtab1}
	\end{table}

	\begin{table}
		\caption{Formulas for calculating the computational complexity and the weights of different spatiotemporal layers. }
		
		\setlength{\tabcolsep}{3pt}
		\begin{tabular}{|p{70pt}|p{50pt}|p{50pt}|p{60pt}|}
			\hline
			\textbf{layer}&
			\textbf{MULs}&
			\textbf{ADDs}&
			\textbf{Weights}
			\\

			\hline
			ConvLIAF&
			\begin{math} (Q + 1) \cdot R \end{math}&
			\begin{math} (Q + 2) \cdot R \end{math}&
			\begin{math} (Q + 1) \cdot L \end{math} \\
			\hline
			ConvLIF&
			\begin{math} R \end{math}&
			\begin{math} (Q + 2) \cdot R \end{math}&
			\begin{math} (Q + 1) \cdot L \end{math} \\
			\hline

			Conv2D (TD)&
			\begin{math} Q \cdot R \end{math}&
			\begin{math} Q \cdot R \end{math}&
			\begin{math} (Q + 1) \cdot L \end{math}  \\
			\hline
			Conv3D&
			\begin{math} U \cdot Q \cdot R \end{math}&
			\begin{math} U \cdot Q \cdot R \end{math}&
			\begin{math} (U \cdot Q + 1) \cdot L \end{math}  \\
			\hline
			ConvLSTM&
			\begin{math} (4 \cdot (Q+I \cdot J \cdot L) + 3) \cdot R  \end{math}&
			\begin{math} (4 \cdot (Q+I \cdot J \cdot L) + 1) \cdot R \end{math}&
			\begin{math} (Q + I \cdot J \cdot L + 1) \cdot 4 \cdot L \end{math}  \\
			\hline
			\multicolumn{4}{p{250pt}}{Note: TD refers to the time-distributed operation, i.e. duplicate over time.}
		\end{tabular}
		\label{evtab2}
	\end{table}
	
\begin{figure*}[!t]
\centerline{\includegraphics[width=500pt]{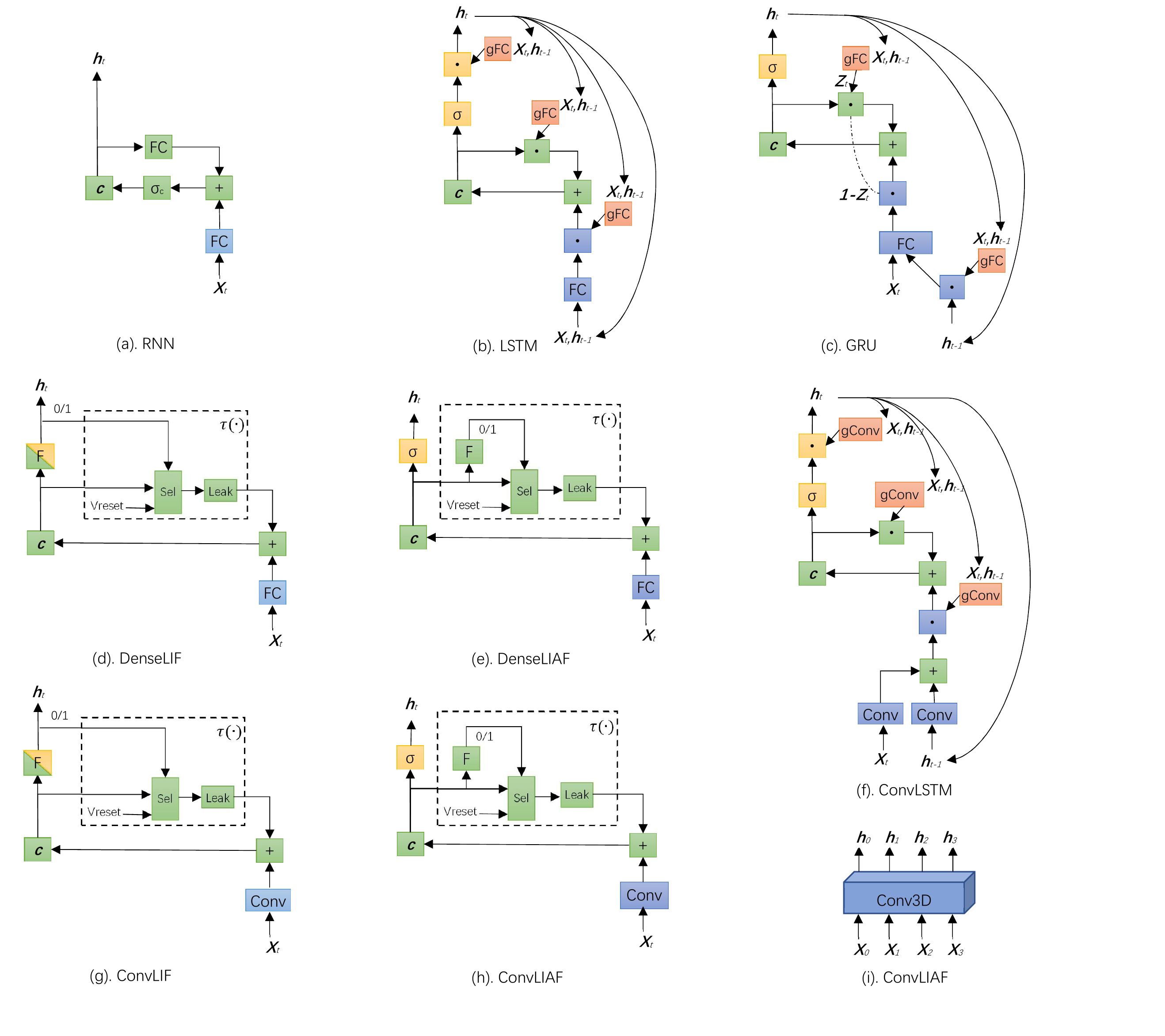}}
\caption{Comparison of LIAF with other temporal and spatiotemporal models. The nodes for hidden state updates are denoted as green nodes, the output nodes are shown in yellow, and the input nodes are colored in blue. The fully connected operations and convolutional opertions for calculating the gates are drawn in orange color. For simplicity, the time step delays for all variables are not drawn. It shows that LIAF and LIF update the hidden state by neuronal dynamics (membrane potential accumulation, leak and action potential generation, potential resetting), without the need of gating. Therefore, several gating (gFC) operators are saved compared to LSTM and GRU, and several convolutional gating operators (gConv) are saved compared to ConvLSTM. Each gFC/gConv requires a large amount of computation and weight storage. In addition, RNN requires additional recurrent weights that do not exist in DenseLIF/DenseLIAF. The main computational and storage overhead in ConvLIAF is the 2D convolution which is much resource-saving than Conv3D. Therefore DenseLIAF and DenseLIF can be viewed as lightweight temporal models compared to RNN, LSTM, and GRU. ConvLIAF and ConvLIF are lightweight spatiotemporal models compared to ConvLSTM and Conv3D.}
\label{fig4}
\end{figure*}

\section{Performance Evaluation}\label{sec:Performance Evaluation}
In this section, we compare the performance and computational cost of LIAF-Net with traditional networks. Firstly, the temporal processing capability of LIAF-Net is evaluated and compared with candidate networks through the Question Answering (QA) tasks. Then we evaluate the spatiotemporal classification accuracy among networks by DVS datasets.

\subsection{Experiment: LIAF-Net on bAbI QA tasks}\label{sec:Experiment: LIAF on bAbI QA tasks}
\begin{figure}[!t]
\centerline{\includegraphics[width=\columnwidth]{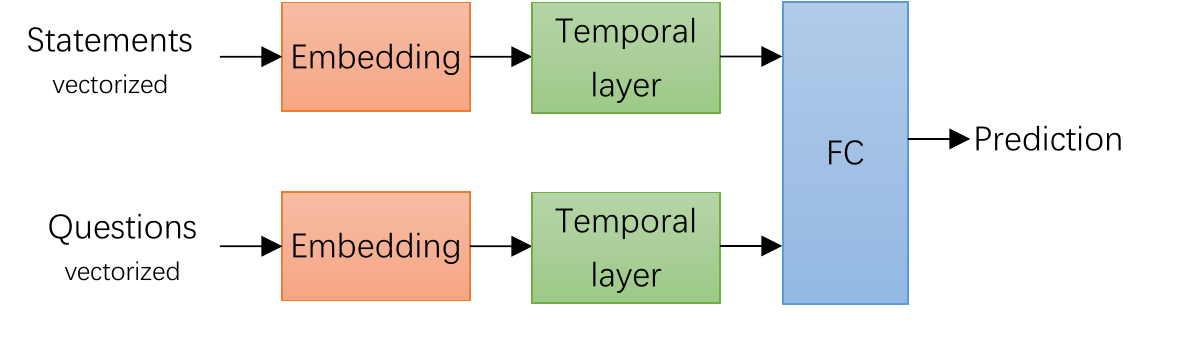}}
\caption{Proposed evaluation network architecture for the bAbI QA tasks. In order to clearly compare the capability of these temporal layers, we keep the network rather simple. Only one single temporal layer selected from LSTM, GRU, LIF, LIAF is applied for processing the temporal information. }
\label{fig5}
\end{figure}

This evaluation measures the performance disparity among the temporal models on simple Natural Language Processing (NLP) tasks. We adopt bAbI dataset \cite{weston2015towards} which is a QA dataset, containing 20 tasks. Each task contains a group of QAs. In each QA there are several statements and questions. A temporal network is required to analyze the statements and memorize the key information (such as a supporting fact, or a relative position relationship), and then answers the question by yes/no or by a word from the statement.
In this work, we introduced a network shown in Fig.\ref{fig5}. For the temporal network, one of DenseLIF/DenseLIAF/GRU/LSTM/RNN is applied. For DenseLIAF, NTR mode is applied with no activation function (expect for QA8 where RELU activation is applied for easy convergence),  for GRU and LSTM, the selected activations are shown in Appendix A. The Statements and Questions are firstly encoded into vectors, then processed by the network. A vocabulary is generated from all the vectors, and the results are denoted by a one-hot coding on the dictionary of the vocabulary. Embedding hidden size (HS) of 50 and Statement/Query HS of 100 is configured. In this experiment, all the temporal layers share the same input/output and training parameters.
We adopt Adam optimizer with categorical cross-entropy as the loss function. For the Equivalent Condition (EC) case, we adopt a learning rate of 0.001 and trained for 60 epochs for all networks. The results are listed in Table \ref{tab1}.

\begin{table*}
\caption{Performance comparison of temporal layers on bAbI QA tasks.}
\label{table*}
\setlength{\tabcolsep}{3pt}
		\begin{tabular}{|p{120pt}|p{70pt}<{\centering}|p{70pt}<{\centering}|p{70pt}<{\centering}|p{70pt}<{\centering}|p{70pt}<{\centering}|}
\hline
\textbf{Task}&
\textbf{RNN EC}&
\textbf{LSTM EC}&
\textbf{GRU EC}&
\textbf{LIF EC (FC)}&
\textbf{LIAF EC (FC)}\\
\hline
QA1 - Single Supporting Fact&
47.8\%&
46.8\%&
49.0\%&
52.7\%&
51.8\%\\
\hline
QA2 - Two Supporting Facts&
27.5\%&
33.6\%&
29.4\%&
27.5\%&
30.7\%\\
\hline
QA3 - Three Supporting Facts&
22.4\%&
22.8\%&
27.4\%&
20.2\%&
20.8\%\\
\hline
QA4 - Two Arg. Relations&
71.3\%&
70.5\%&
52.9\%&
31.1\% (72.2\%)&
42.8\% (73.5\%)\\
\hline
QA5 - Three Arg. Relations&
39.3\%&
73.5\%&
73.0\%&
32.2\%&
68.5\%\\
\hline
QA6 - yes/No Questions&
49.5\%&
50.8\%&
51.2\%&
52.6\%&
50.2\%\\
\hline
QA7 - Counting&
79.3\%&
79.0\%&
76.4\%&
48.8\% (72.8\%)&
75.2\%\\
\hline
QA8 - Lists/Sets&
53.8\%&
75.7\%&
73.4\%&
33.6\%&
41.3\% (73.3\%)\\
\hline
QA9 - Simple Negation&
61.0\%&
63.8\%&
62.4\%&
61.2\%&
56.2\% (60.1\%)\\
\hline
QA10 - Indefinite Knowledge&
45.1\%&
46.8\%&
46.7\%&
46.4\%&
48.2\%\\
\hline
QA11 - Basic Coreference&
69.7\%&
65.5\%&
67.1\%&
75.1\%&
75.1\%\\
\hline
QA12 - Conjunction&
64.8\%&
64.5\%&
62.7\%&
77.2\%&
77.2\%\\
\hline
QA13 - Compound Coreference&
93.6\%&
92.0\%&
91.4\%&
94.4\%&
94.4\%\\
\hline
QA14 - Time Reasoning&
27.4\%&
38.4\%&
39.7\%&
29.1\%(34.1\%)&
29.1\%\\
\hline
QA15 - Basic Deduction&
45.3\%&
24.2\%&
45.9\%&
25.7\%&
23.3\% (38.4\%)\\
\hline
QA16 - Basic Induction&
44.9\%&
46.7\%&
44.4\%&
45.4\%&
45.3\% \\
\hline
QA17 - Positional Reasoning&
48.0\%&
48.0\%&
48.8\%&
49.6\%&
51.6\%\\
\hline
QA18 - Size Reasoning&
90.4\%&
90.3\%&
91.2\%&
90.6\%&
90.1\%\\
\hline
QA19 - Path Finding&
10.2\%&
9.6\%&
8.8\%&
8.6\%&
8.1\%\\
\hline
QA20 - Agent's Motivations&
93.4\%&
97.1\%&
96.7\%&
92.6\%&
93.5\%\\
\hline
\textbf{Average performance over all tasks}&
\textbf{54.2\%}&
\textbf{57.0\%}&
\textbf{56.9\%}&
\textbf{49.7\%} (\textbf{53.2\%})&
\textbf{53.7\%} (\textbf{57.8\%})\\
\hline
\multicolumn{6}{p{500pt}}{Note: EC means Equivalent Learning Condition (all networks are configured with lr=0.001, and train in 60 epochs), FC indicates Finetune Learning Condition (lr = 0.0005, 200 epochs for QA8, lr = 0.0005, 500 epochs for LIAF-QA4/9/15 LIF-QA4/7/14, other tasks are the same with EC).}\\
\end{tabular}
\label{tab1}
\end{table*}

We evaluated the average scores of all tasks with the temporal models. It shows that LSTM and GRU based network reach similar performance on accuracy. LIAF (EC) reaches slightly weak performance (-3.3\%) to LSTM/GRU and -0.5\% to RNN. In addition, LIAF still obtains an obvious performance gain to LIF (3.9\%) with the same number of weight parameters. Note that RNN contains an additional internal FC which resulting in much higher storage and computation than LIAF.

It shows in Table \ref{tab1} that LIAF performs weaker than LSTM especially in task QA4 and QA8, which motivates us to find out the reason. Further experiments reveal that if we train LIAF with more epochs and lower learning rate (0.0005), it achieves dramatic performance gain, shown in LIAF (FC). With finetuning, LIF achieves higher accuracy on QA4/7/14 shown in LIF (FC). We also found that under this configuration, LSTM and GRU obtain limited performance gain. In conclusion, after finetuning the configuration, LIAF obtains comparable performance to LSTM and GRU, and still outperforms LIF for 4.6\%.

We further investigated the characteristic of these tasks, the results revealed that for limited support facts and simple reasoning logic, LIAF / LIF (EC case) may perform better than LSTM / GRU, which happens in QA11 and QA12. On the contrary, for complicated relationships and multi-hop reasoning, LSTM / GRU may perform better than LIAF / LIF (EC case) revealed in QA8 and QA14, which may be caused by the absence of gating in LIAF / LIF.

We revealed that LIAF-Net can be treated as a lightweight temporal network model, therefore we also measured the number of parameters and computational operations in these networks. The results are shown in Table \ref{compbabi} which reveals that with the same number of hidden units (size = 100), a LIAF layer consumes 63.5\% / 90.8\% / 87.7\% fewer weight parameters over an RNN / LSTM / GRU layer respectively. In addition, LIAF also has lower computational overhead. A LIAF layer has 91.5\% less computational overhead than an LSTM layer and 88.8\% than a GRU layer.

	\begin{table}
		\caption{Efficiency and cost comparison for the candidate temporal models on bAbI dataset.}
		\setlength{\tabcolsep}{3pt}
		\begin{tabular}{|p{40pt}|p{40pt}|p{40pt}|p{40pt}|p{40pt}|}
			\hline
			\textbf{Temporal layer}&
			\textbf{ADDs (layer)}&
			\textbf{MULs (layer)}&
			\textbf{Weights (layer)}&
			\textbf{Weights (network)}
			\\
			\hline
			DenseLIAF&
			\begin{math} 3.4 \times {10}^{5} \end{math}&
			\begin{math} 3.3 \times {10}^{5} \end{math}&
			\begin{math} 5.5 \times {10}^{3} \end{math} &
			\begin{math} 1.7 \times {10}^{4} \end{math}  \\
			\hline
			DenseLIF&
			\begin{math} 3.4 \times {10}^{5} \end{math}&
			\begin{math} 6.6 \times {10}^{3} \end{math}&
			\begin{math} 5.5 \times {10}^{3} \end{math} &
			\begin{math} 1.7 \times {10}^{4} \end{math}  \\
			\hline
			RNN&
			\begin{math} 9.9 \times {10}^{5} \end{math}&
			\begin{math} 9.9 \times {10}^{5} \end{math}&
			\begin{math} 1.5 \times {10}^{4} \end{math} &
			\begin{math} 3.6 \times {10}^{4} \end{math}  \\
			\hline
			GRU&
			\begin{math} 2.9 \times {10}^{6} \end{math}&
			\begin{math} 3.0 \times {10}^{6} \end{math}&
			\begin{math} 4.5 \times {10}^{4} \end{math} &
			\begin{math} 9.7 \times {10}^{4} \end{math}  \\
			\hline
			LSTM&
			\begin{math} 3.9 \times {10}^{6} \end{math}&
			\begin{math} 4.0 \times {10}^{6} \end{math}&
			\begin{math} 6.0 \times {10}^{4} \end{math} &
			\begin{math} 1.2 \times {10}^{5} \end{math}  \\
			\hline
		\end{tabular}
		\label{compbabi}
	\end{table}


\subsection{Experiment: MNIST-DVS and CIFAR10-DVS}\label{sec:Experiment: Spatiotemporal DVS Sensing}
\subsubsection{Datasets and preparation}\label{sec:Datasets and Preparation}
For spatiotemporal tasks, two neuromorphic datasets, MNIST-DVS \cite{serrano2015poker} and CIFAR10-DVS \cite{li2017cifar10}, are used to verify the performance of LIAF-Net.  Compared to image-based datasets, these event-based datasets contain abundant temporal information. The captured event-based data is typically spike train, where each spike is triggered by a light intensity change at each pixel. Spike is represented by a quad such as ($x, y, ts, pol$), where $x$ and $y$ are the spatial coordinates of the spike, $t s$ is the time-stamp of the event (unit of 1$\mu s$), and $pol$ represents the type of light intensity change (lighten to 1 or darken to -1).

For data pre-processing, we generated a single event-frame by accumulating the spike train within every 5ms. Specifically, we set up different channels for diverse light intensity changes, and adjacent $T$ event frames in chronological order are used, then a sample is derived in size ($T$, 128, 128, 2). In this case, a larger $T$ indicates that the sample has more temporal information. We set $T=20$ for a MNIST-DVS sample and $T=10$ for a CIFAR10-DVS sample. For CIFAR10-DVS we used the full DVS 128$\times$128 pixel resolution, while for MNIST-DVS we cropped the event-frame from 128$\times$128 to the size of 40$\times$40 according to the position of the handwritten digits. MNIST-DVS has three scales (scale4, scale8, scale16), and we chose scale8 by default.

\subsubsection{Network structure}\label{sec:Network structure}
\begin{figure*}[!t]
\centerline{\includegraphics[width=500pt]{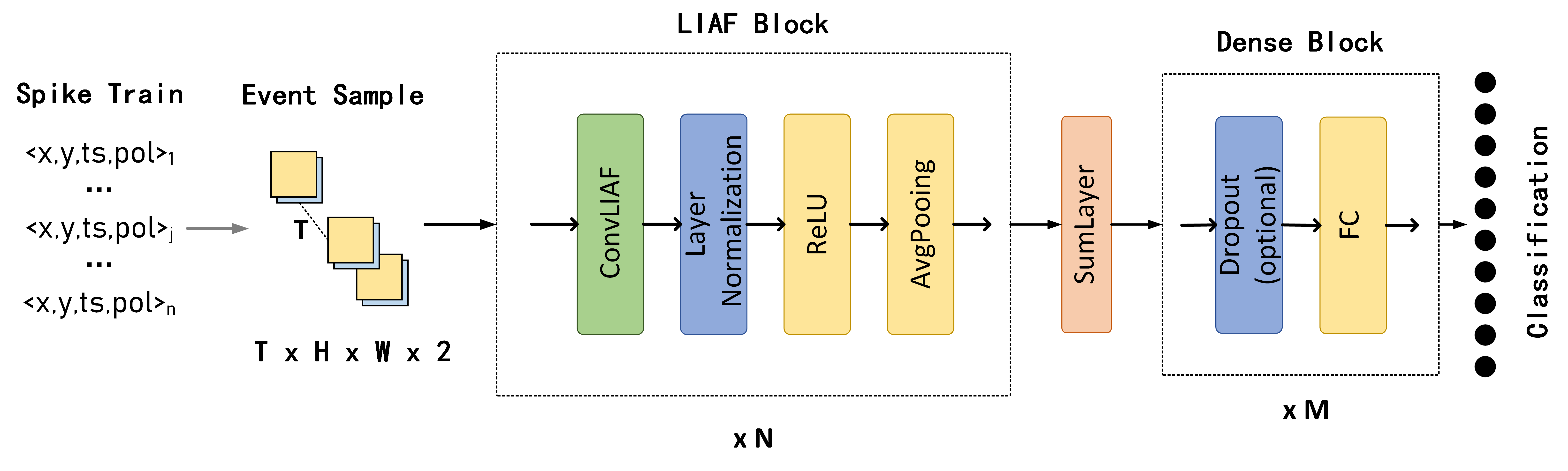}}
\caption{Illustration of network structure used for classification tasks. Networks for different tasks have different N and M. N denotes that N LIAF Blocks are connected end to end. M means that M Dense Blocks are connected end to end. Both MNIST-DVS and CIFAR10-DVS experiments share the same network structure which indicates the generality of the structure.}
\label{fig7}
\end{figure*}

We proposed a flexible network structure for these classification tasks. The proposed network structures are VGG-like \cite{simonyan2014very} illustrated in Fig.\ref{fig7}, where the letter N denotes that the LIAF Blocks are connected end to end for N times and similarly the letter M represents that the Dense Blocks are connected end to end for M times. LIAF Block in Fig.\ref{fig7} has a sequential structure, consisting of ConvLIAF, TD-layer-normalization, TD-activation (ReLU) and TD-AvgPooling, where TD refers to the time-distributed operation. Dense Block consists of a dropout layer and a dense layer. In particular, the kernel size of layer ConvLIAF is set to (3, 3), and the padding is always 1. Among LIAF configurations, the one with TR mode activation and Channel-Sharing is adopted, with the activation $Act(x)=x$. $\bm{V}_{th}$, $\bm{V}_{reset}$, $\bm{\alpha}$ and $\bm{\beta}$ are all trained by BPTT. The parameter settings are listed in Table \ref{tab4}. For LIF-SNN, the LIAF Block is replaced by LIF Block consisting of sequentially connected layers including TD-Conv2D, TD-layer-normalization, TD-AvgPooling and DirectLIF.

Since Layer normalization \cite{ba2016layer} is very effective at stabilizing the hidden state dynamics in recurrent networks, we adopt it to accelerate the inference and training of the network and prevent overfitting. It is noteworthy that the feature still has temporal dimension after passing through the N LIAF blocks. For integrating the temporal information, we designed Sumlayer, which reduces the dimension of the activation neuron shape from $(T, H, W, L)$ to $(H, W, L)$ by element-wise addition of the activations over all time steps and divided by $T$. Finally, the network ends with a softmax layer with 10 units for classification.
\begin{table}
\caption{Parameter settings for MNIST-DVS and CIFAR10-DVS .}
\setlength{\tabcolsep}{3pt}
		\begin{tabular}{|p{50pt}|p{10pt}|p{10pt}|p{40pt}<{\centering}|p{40pt}<{\centering}|p{50pt}|}
\hline
\textbf{Task}&
\textbf{N}&
\textbf{M} &
\textbf{Block name} &
\textbf{Filters or units} &
\textbf{Parameter} \\
\hline
\multirow{5} * {MNIST-DVS}
& ~ & ~ & LIAF & 32& Pooling (2, 2) \\
\cline{4-6}
~ & ~ & ~ & LIAF & 64 & Pooling (2, 2) \\
\cline{4-6}
~ & 3 & 2 & LIAF & 128 & Pooling (2, 2) \\
\cline{4-6}
~ & ~ & ~ & Dense & 512 & - \\
\cline{4-6}
~ & ~ & ~ & Dense & 128 & - \\
\hline
\multirow{5} * {CIFAR10-DVS}
& ~ & ~ & LIAF & 32& Pooling (2, 2) \\
\cline{4-6}
~ & ~ & ~ & LIAF & 64 & Pooling (2, 2) \\
\cline{4-6}
~ & 5 & 1 & LIAF & 128 & Pooling (2, 2) \\
\cline{4-6}
~ & ~ & ~ & LIAF & 256 & Pooling (2, 2) \\
\cline{4-6}
~ & ~ & ~ & LIAF & 512 & Pooling (4, 4) \\
\cline{4-6}
~ & ~ & ~ & Dense & 512 & - \\
\hline
\end{tabular}
\label{tab4}
\end{table}

\subsubsection{Training setup}\label{sec:Training Setup}
We trained the network shown in Fig.\ref{fig7} and Table \ref{tab4} on MNIST-DVS and CIFAR10-DVS datasets respectively using the optimizer Adam and learning rate of 5e-5. Necessarily, we used a learning rate fine-tuning strategy, which is to divide the learning rate by 5 when the loss of validation has stopped improving for the latest 5 epochs.


\begin{table}
\caption{Comparison with state-of-the-art results on MNIST-DVS and CIFAR10-DVS. }
\label{table cfg}
\setlength{\tabcolsep}{3pt}
\begin{tabular}{|m{70pt} |m{70pt}|p{40pt}<{\centering}|p{40pt}<{\centering}|}
\hline
\textbf{Proposals} &
\textbf{Methods}&
\textbf{MNIST-DVS} &
\textbf{CIFAR10-DVS} \\
\hline
Zhao et al. 2014 \cite{zhao2014feedforward}&
Convolutional SNN&
88.14\%&
- \\
\hline
Stromatias et al. 2017 \cite{stromatias2017event}&
Composite system&
97.95\%&
- \\
\hline
Lagorce et al., 2017 \cite{lagorce2016hots}&
HOTS&
-&
27.10\% \\
\hline
Shi et al., 2018 \cite{shi2018exploiting}&
Lightweight Statistical&
78.08\%&
31.20\% \\
\hline
Paulun et al.2018 \cite{paulun2018retinotopic}&
NeuCube&
92.03\%&
- \\
\hline
Cannici et al., 2018 \cite{cannici2019attention}&
Attention Mechanisms&
-&
44.10\% \\
\hline
Sironi et al., 2018 \cite{sironi2018hats}&
HATS&
98.40\%&
52.40\% \\
\hline
Wu et al., 2019 \cite{wu2019direct}&
-&
-&
60.50\% \\
\hline
This work&
LIF&
97.51\%&
\textbf{63.53\%} \\
\hline
This work&
LIAF&
\textbf{99.13\%}&
\textbf{70.40\%} \\
\hline
\end{tabular}
\label{tab5}
\end{table}
\subsubsection{Performance analysis}
\label{sec:Performance Analysis}
\begin{figure*}[!t]
\centerline{\includegraphics[width=450pt]{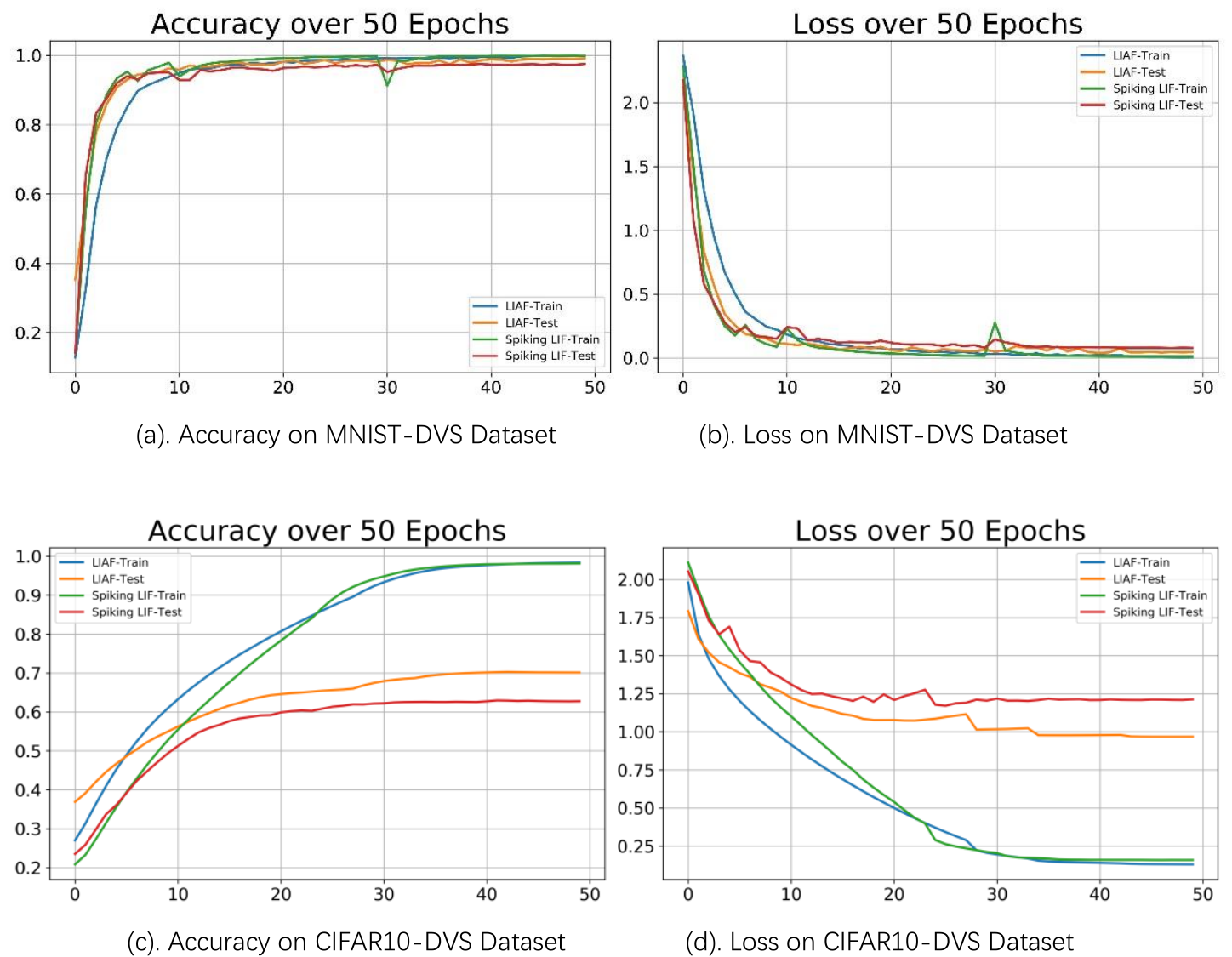}}
\caption{The comparison of LIAF-Net and LIF-SNN in terms of accuracy and loss for all the training epochs on the MNIST-DVS Dataset and CIFAR10-DVS Dataset.}
\label{fig8}
\end{figure*}

We tested the proposed networks on corresponding test datasets occupying 20\% of the entire dataset and compare the results with existing state-of-the-art methods summarized in Table \ref{tab5}. On one hand, we have achieved extremely high accuracy on both the MNIST-DVS and CIFAR10-DVS datasets, in special on the CIFAR10-DVS dataset, we achieved the best accuracy, 70.4\%, which is 9.9\% higher than the best result of the previous work. On the other hand, by comparing the results of LIAF-Net and LIF-SNN, we have demonstrated that LIAF-Net can better analyze event-based data than LIF-SNN. It is shown in Fig. \ref{fig8} that the curve for LIAF-Net are smoother than the LIF-SNN and receives less overfitting, which may be caused by the fact that the differential function over continuous activation in LIAF is smoother than the Heaviside step function for the threshold firing in LIF.

	\begin{table}
		\caption{Performance and resource consumption comparison among spatiotemporal models on CIFAR10-DVS.}
		\setlength{\tabcolsep}{2pt}
		\begin{tabular}{|p{45pt}|p{40pt}|p{40pt}|p{40pt}|p{30pt}|p{30pt}|}
			\hline
			\textbf{Network}&
			\textbf{MULs} &
			\textbf{ADDs} &
			\textbf{Weights*} &
			\textbf{Train acc.} &
			\textbf{Test acc.} \\
			\hline
			Conv2D&
			\begin{math} 3.3 \times {10}^{9} \end{math}&
			\begin{math} 3.8 \times {10}^{9} \end{math}&
			\begin{math} 1.5 \times {10}^{6} \end{math}&
			97.8\%&
			67.8\%  \\
			\hline
			Conv3D&
			\begin{math} 9.5 \times {10}^{9} \end{math}&
			\begin{math} 1.0 \times {10}^{10} \end{math}&
			\begin{math} 4.7 \times {10}^{6} \end{math}&
			99.8\%&
			71.7\%  \\
			\hline
			ConvLSTM&
			\begin{math} 4.2 \times {10}^{10} \end{math}&
			\begin{math} 4.3 \times {10}^{10} \end{math}&
			\begin{math} 1.8 \times {10}^{7} \end{math}&
			100\%&
			70.8\%  \\
			\hline
			ConvLIAF&
			\begin{math} 3.3 \times {10}^{9} \end{math}&
			\begin{math} 3.8 \times {10}^{9} \end{math}&
			\begin{math} 1.5 \times {10}^{6} \end{math}&
			98.4\%&
			70.4\%  \\
			\hline
			ConvLIF&
			\begin{math} 2.1 \times {10}^{8} \end{math}&
			\begin{math} 3.8 \times {10}^{9} \end{math}&
			\begin{math} 1.5 \times {10}^{6} \end{math}&
			97.5\%&
			63.5\%  \\
			\hline
			\multicolumn{6}{p{250pt}}{Note*: LIF and LIAF contain additional trainable parameters including $\bm{V}_{reset}$, $\bm{V}_{th}$, $\bm{\alpha}$, $\bm{\beta}$, which are counted in weights.}
		\end{tabular}
		
		\label{tabdvs}
	\end{table}
	

\begin{figure*}[!t]
\centerline{\includegraphics[width=500pt]{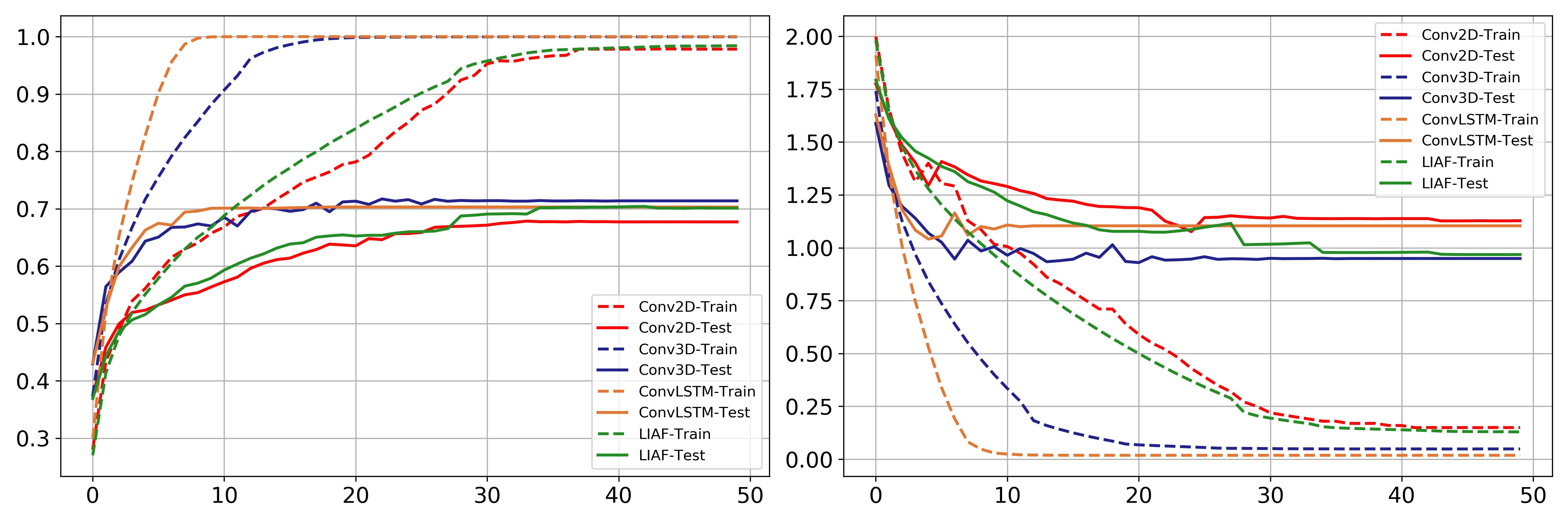}}
\caption{The comparison of LIAF-Net and other traditional spatiotemporal networks in terms of accuracy (left) and loss (right) for all the training epochs on the CIFAR10-DVS Dataset.}
\label{fig9}
\end{figure*}

\subsubsection{Evaluating the efficiency of LIAF-Net with other spatiotemporal networks}\label{sec:Evaluating the efficiency of LIAF with other spatiotemporal models}
We further compared the performance and the costs of LIAF-Net with traditional spatiotemporal networks built by ConvLIAF, time-distributed Conv2D, Conv3D, and ConvLSTM on the CIFAR10-DVS dataset. The network structure is shown in Fig. \ref{fig7} and parameter settings of all networks are listed in Table \ref{tab4}. The spatiotemporal layers are compared on the basis that all models share the same input and output. We set the kernel size of the Conv3D to (3, 3, 3) similar to (3, 3) in Conv2D. All models are trained with Adam optimizer and binary cross-entropy loss function. The initial learning rate are slightly different (settings for Conv2D, Conv3D, and ConvLSTM are 5e-4, 5e-5, 5e-5 respectively) for achieving better performance.
For promoting the convergence of ConvLSTM, the learning rate fine-tuning strategy is adopted, in which the reduction factor is 0.2, but the number of monitoring epochs is reduced from 5 to 2.

The result shown in Table \ref{tabdvs} indicates that LIAF-Net receives the best performance/cost balance compared with other spatiotemporal models. The accuracy and loss curves of all models over 50 training epochs are illustrated in Fig. \ref{fig8}. We observe that although Conv3D and ConvLSTM based networks have higher training accuracy than LIAF-Net, both suffer severe overfitting, which may due to the redundancy of weights. Besides, although Conv2D network and LIAF-Net converge at a similar speed, the Conv2D based network performance is still lower than LIAF-Net after 30 epochs, which may be caused by the absent of temporal processing capability in the Conv2D based network.

In another aspect, LIAF-Net can still achieve better performance meanwhile saving storage and computational workload. The results show in Table \ref{tabdvs} verify that LIAF-Net achieves approximate performance with 68\% and 92\% fewer weight parameters than Conv3D based network and ConvLSTM based network respectively. Table \ref{tabdvs} also shows that the computational overhead of ConvLIAF is 64\% and 92\% less than that of Conv3D and ConvLSTM, respectively under the same activation tensor size.

\subsection{Experiment: DVS128 gesture recognition}\label{sec:DVS-gesture}

\begin{figure*}[!t]
\centerline{\includegraphics[width=500pt]{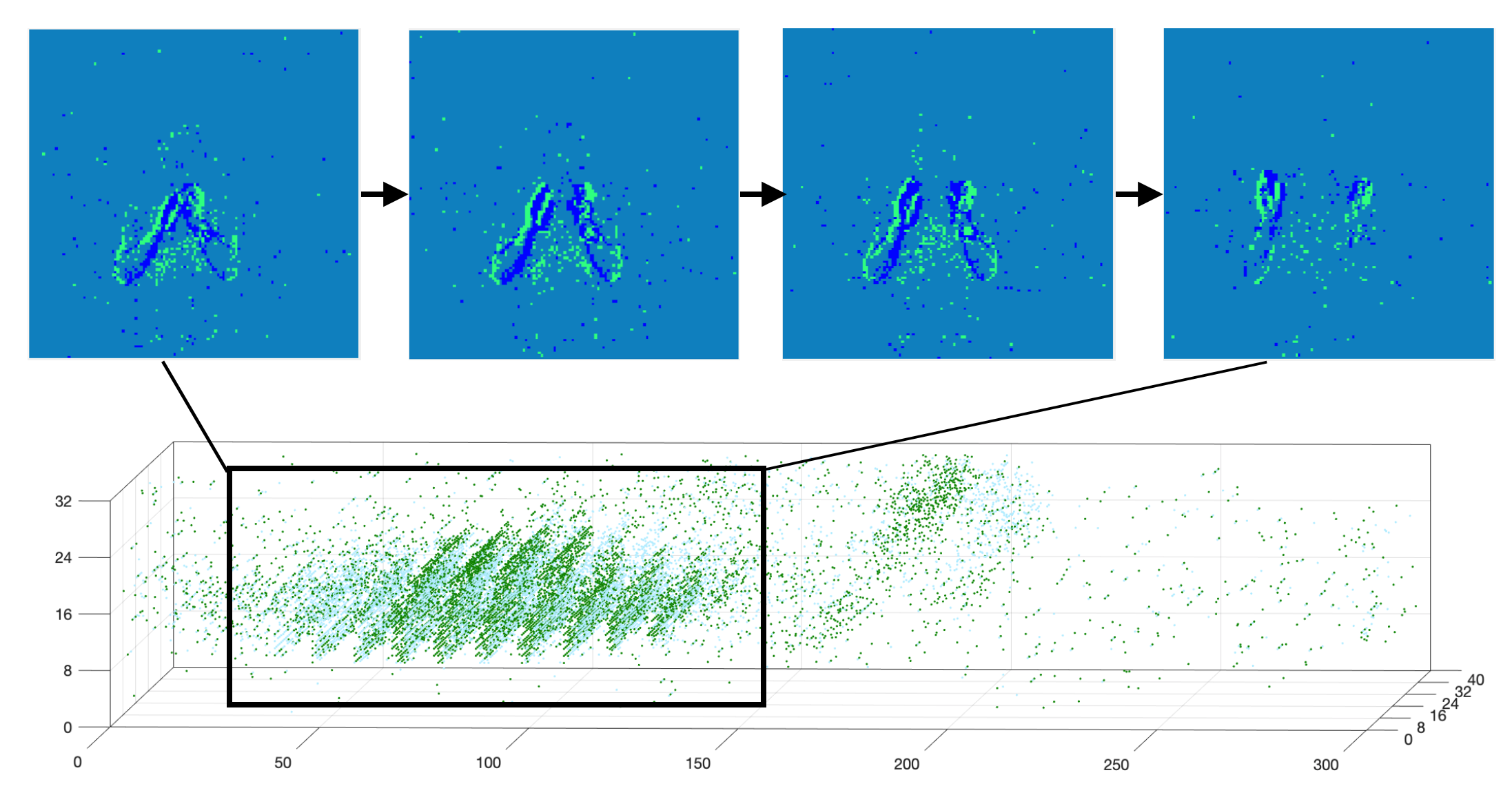}}
\caption{The event stream of a hand-clap gesture is shown in the time-width-height space, where green dots denote that polarity of an event is positive and blue ones denote negative polarity. The camera only records the changing pixel so we can only recognize the edge of the moving objects. Four frames generated by accumulating a period of events are shown above, which tells that a person is clapping his hands.\cite{IBMwebsite}}
\label{figxi}
\end{figure*}


\subsubsection{Dataset and preparation}\label{sec:Datasets and Preparation2}
DVS128 Gesture Dataset\cite{Amir_2017_CVPR} is directly recorded by DVS camera from real scene. This dataset contains 1,342 instances of a set of 11 hand and arm gestures under three different illumination conditions. The data sample is also event-based, and the event is represented as a quad ($ts,x,y,pol$). It has a raw spatial pixel resolution of $128\times 128$ and the $1/4$ down-sampled resolution is applied, to save the graphics memory. To use the network structure for training, we generated event-frames with the size of $(32, 32)$ by accumulating the spike trains within every 25ms. Each frame is then expanded into two channels according to whether the illumination in each pixel is weakened or strengthened. A sample of hand-clapping gesture is visualized in Fig. \ref{figxi}. Finally, multiple adjacent event frames are stacked in a chronological order to obtain a sample with the dimension $(2, 32, 32, T)$. We chose $T=60$ frames for the three networks we used for comparison. After the pre-processing, the data is organized with the form of (Batchsize, Channels, Width, Height, $T$) = $(36, 2, 32, 32, 60)$. Notes that in DVS128 Gesture Dataset there are two testing tasks including classification of 10 classes and 11 classes. We chose the latter one for the experiment.

\subsubsection{Network structure and training setup}\label{sec:Network structure2}

The network structure used here is slightly different from the one in Fig. \ref{fig7}. Here all layers in the network are built using LIAF model while in Fig. \ref{fig7} pooling layers and dense layers are still classical artificial neuron models. LIAF-Net is mainly built with ConvLIAF blocks and DenseLIAF blocks. A ConvLIAF block contains layers including Conv2D, BatchNorm, activation, PoolingLIAF (optional). DenseLIAF block contains layers including Fully connection, BatchNorm, and activation. PoolingLIAF is formed by a pooling layer followed by a DirectLIAF layer. The parameter N now denotes that the N ConvLIAF Blocks are connected end to end, similarly, the letter M represents that M DenseLIAF Blocks are connected end to end. The network settings are listed in Table \ref{tabx}.

\begin{table}
\caption{Parameter settings for the DVS128 gesture recognition experiment.}
\setlength{\tabcolsep}{3pt}
		\begin{tabular}{|p{40pt}|p{10pt}|p{10pt}|p{40pt}|p{30pt}<{\centering}|p{70pt}|}
\hline
\textbf{Task}&
\textbf{N}&
\textbf{M} &
\textbf{Block name} &
\textbf{Output filters or units} &
\textbf{Parameter} \\
\hline
& ~ & ~ & ConvLIAF & 64 & No pooling\\
\cline{4-6}
~ & ~ & ~ & ConvLIAF & 128 &  PoolingLIAF (2, 2) \\
\cline{4-6}
DVS- & 3 & 2 & ConvLIAF & 128 & PoolingLIAF (2, 2) \\
\cline{4-6}
Gesture & ~ & ~ & DenseLIAF & 256 & - \\
\cline{4-6}
~ & ~ & ~ & DenseLIAF & 11 & - \\
\hline
\end{tabular}
\label{tabx}
\end{table}

There are also some parameters specific to LIAF, wherein NTR and All-Sharing configurations are adopted. We set $V_{th}=0.5$ in equation (4), $\alpha=0.3$ and $\beta=0$ in equation (6). The scaled exponential linear unit (SELU)\cite{NIPS2017_6698} is applied as the activation function in DirectLIAF, which induces a self-normalizing property. For LIF-SNN, all LIAF layers in the network are replaced by LIF layers. For ConvLSTM network, all LIAF layers are replaced by ConvLSTM layers. All convolutional kernels in these networks are in the size of $3 \times 3$.

We trained the three networks via Adam optimizer with a  learning rate of 1e-4 and the weight decay of 1e-4. Necessarily, we used learning rate fine-tuning strategies for each experiment.

\subsubsection{Performance analysis}\label{sec:Performance Analysis2}

We tested the proposed networks on the corresponding test set which includes 288 instances of arm and hand gestures. The test set accuracy result shows that we have achieved a new record on DVS gesture using LIAF-Net with an accuracy of 97.56\%, which is 3.46\% higher than the result of the LIF-SNN and also 3.46\% higher than the result of ConvLSTM based network.
We also compared our solution with related works and the top 1 accuracy of all solutions is listed in Table \ref{dvsgcmp}. It also shows that analog networks achieve better accuracy than spiking networks on average, and our proposal achieves the best accuracy among them, which reveals the spatiotemporal processing capability of LIAF.

In addition, the efficiency advantage of LIAF-Net in terms of less convolutional parameters and computational overhead required than ConvLSTM based network is illustrated in Table \ref{evgesture}. We only list the computational overhead and weights of all convolutional layers in the network to more purely reflect the difference in efficiency between LIAF and LSTM, which reveals that ConvLIAF saves 91.6\% of the computation overhead and 90.0\% of the storage compared to ConvLSTM.


	\begin{table}
		\caption{Comparison with the accuracy results and resource consumption on the DVS128 Gesture Dataset.}
		\setlength{\tabcolsep}{2pt}
		\begin{tabular}{|p{40pt}|p{35pt}|p{35pt}|p{35pt}|p{35pt}|p{35pt}|}
			\hline
			\textbf{Network} &
			\textbf{Test acc.} &
			\textbf{Neurons} &
			\textbf{ADDs} &
			\textbf{MULs} &
			\textbf{Weights}
			\\
			\hline
			ConvLIAF&
			97.56\%&
			\begin{math} 2.3 \times {10}^{5} \end{math}&
			\begin{math} 6.8 \times {10}^{9} \end{math}&
			\begin{math} 6.8 \times {10}^{9} \end{math}&
			\begin{math} 2.2 \times {10}^{5} \end{math}  \\
			\hline
			ConvLIF&
			94.10\%&
			\begin{math} 2.3 \times {10}^{5} \end{math}&
			\begin{math} 6.8 \times {10}^{9} \end{math}&
			\begin{math} 1.3 \times {10}^{7} \end{math}&
			\begin{math} 2.2 \times {10}^{5} \end{math} \\
			\hline
			ConvLSTM&
			94.10\%&
			\begin{math} 2.3 \times {10}^{5} \end{math}&
			\begin{math} 8.1 \times {10}^{10} \end{math}&
			\begin{math} 8.1 \times {10}^{10} \end{math}&
			\begin{math} 2.2 \times {10}^{6} \end{math}  \\
			\hline
		\end{tabular}
		\label{evgesture}
	\end{table}

\begin{table}
\caption{Top1 Accuray of Solutions for the DVS128 Gesture Dataset (11 classes). }
\label{dvsgcmp}
    \centering
    \begin{tabular}{|p{70pt}<{\centering}|p{80pt}<{\centering}|p{25pt}<{\centering}|p{25pt}<{\centering}|}
    \hline
        \textbf{Proposals} & \textbf{Methods} & \textbf{Act.} & \textbf{Acc.}\\
        \hline
        Massa, et al.,  2020 \cite{massa2020efficient}& SNN converted from CNN on Loihi & spike & 89.64\% \\
        \hline
        Amir et al., 2017\cite{amir2017low} & CNN on TrueNorth & spike & 94.59\% \\
        \hline
        Kugele et al., 2020\cite{kugele2020efficient}& SNN Converted form ANN & spike & 95.56\% \\
        \hline
        \textbf{This work} & SNN (ConvLIF) & spike & 94.10\% \\
        \hline
        Khoei et al., SpArNet 2020\cite{khoei2020sparnet}& Converted CNN & analog & 95.1\% \\ \hline
        Wang et.al. 2019\cite{wang2019space}& PointNet++ &analog & 95.32\% \\ \hline
        Bi et al., 2019\cite{bi2019graph}& Residual graph CNN + Res. 3D & analog & 97.2\% \\

        \hline
        \textbf{This work}& LIAF-Net (ConvLIAF) & analog & \textbf{97.56\%} \\
        \hline
    \end{tabular}
\end{table}

\section{Discussions}\label{sec:Discussions}
\subsection{Rethinking of the bio-plausibility of LIAF}
There is also evidence on the existence of analog action potentials (spikes) in vivo  \cite{gidon2020dendritic}, where a multi-valued calcium-mediated dendritic action potentials is discovered in the L2/3 pyramidal neurons of the human cerebral cortex. The amplitude varies with different stimuli levels. It is possible to model such dendritic behavior using a LIAF-like model whose action potentials are analog. Such discovery also indicates the bio-plausibility of LIAF.

\subsection{Artificial neuron models using analog value}
Beyond the neuron models in ANNs where analog value is the basic format for the activations, there are more spiking neuron models using analog value. In several Spiking Neural Network models, analog values are derived from the traces of the spike trains \cite{kulkarni2017learning} \cite{soures2019deep} which also reveals biological consistency \cite{dayan2001theoretical}. In another aspect, for modeling a complicated dendrite, multi-compartment model \cite{dayan2001theoretical} is necessary, where the current is transmitted from one compartment to another through a conductance. The current between these compartments is an analog value other than spike. In addition, it is admissible to accept analog-valued batch normalization or layer normalization operations in a bio-plausible neural network since it can be viewed as a type of intrinsic homeostasis \cite{tang2019bridging} of a neuron cell. To reduce the computational overhead, SpArNet\cite{khoei2020sparnet} is proposed where network layers communicate with analog spikes. Only above threshold activations are transmitted for dramatically reducing the synaptic operations. Different from LIAF-Net which is trained on spatiotemporal data directly, the network is converted from pretrained CNN.

\subsection{From LIAF to SNN}
It is possible to build pure SNN through LIAF by introducing rate or temporal coding. The model in previous sections assume that the neuron dynamic of LIAF is based on a basic unit called ‘time step’, and each time step has a single analog value on either dendrite (input) or axon (output). If we introduce sub temporal intervals for a time step, more coding schemes can be realized. Then an analog value can be represented by a train of spikes within the time step, wherein the firing rate or the firing time represents the analog value. Although many sub-intervals are required for approximating an accurate analog value,  limited sub-intervals can represent a quantized analog activation.
It is interesting that LIAF can also be used as an intermediate representation for converting rate and temporal coding spike trains from one to another.

\section{Conclusion}\label{sec:Conclusion}
In this work, we proposed a Leaky Integrate and Analog Fire (LIAF) neuron model and several LIAF-Nets built on it for efficient spatiotemporal processing. LIAF-Net makes it  easier for the back propagation to be applied and maintains the spatiotemporal processing capability through the LIF model dynamics. It also benefits from ANN layers, training techniques, and network building frameworks. By introducing LIAF, SNNs and ANNs can communicate with each other more easily without coding format conversion. Therefore   LIAF-Net   provides a framework that can be used to friendly build large scale networks. In the performance evaluation section, it is demonstrated that LIAF-Net can be applied for solving various real temporal and spatiotemporal tasks with higher accuracy while consumes lower computational and storage costs compared with existing traditional networks.

A common viewpoint is that current ANN is the 2nd generation neural network, and SNN is the 3rd generation \cite{roy2019towards} because of its event-driven advantages and bio-plausibility. However, SNN encounters training difficulty with back propagation (BP) algorithm due to the spiking format, which makes the evolution not easy. Since LIAF introduces a more flexible activation function to LIF, and LIF is a special case of LIAF, LIAF can be viewed as a 3.5th generation evolving technology for LIF SNN upgrading.

\bibliographystyle{./IEEEtran}
\bibliography{liaf}

\section{Appendix}
\subsection{Formal definitions for the referenced models}
\subsubsection{RNN}
We use Elman RNN \cite{ wiki:RNN} as a reference, following

\begin{equation}
\begin{aligned}
c_t&=\sigma_c(W_cx_t+U_cc_{t-1}+b_c); \\
h_t&=\sigma_h(W_hc_t+b_h).
\end{aligned}
\end{equation}

where $x_{t}$ is the input vector;
$c_{t}$ is the hidden state vector;
$h_{t}$ is the output vector;
$W_c, W_h, U_c, b_c, b_h$ are trainable parameters;
$\sigma_{c}$ and $\sigma_h$ are activation functions, we use $\sigma_{c}=tanh$ by default.

Note that we exclude the second equation from our RNN model since it can be realized by a fully connected layer outside, and the remaining part is more similar to LIF/LIAF.

\subsubsection{LSTM and ConvLSTM}

\begin{equation}
\begin{aligned}
f_t&=\sigma_g(W_f * x_t+U_f * h_{t-1}+b_f); \\
i_t&=\sigma_g(W_i * x_t+U_i * h_{t-1}+b_i); \\
o_t&=\sigma_g(W_o * x_t+U_o * h_{t-1}+b_o); \\
\tilde{c}_t&=\sigma_c(W_c * x_t+U_c * h_{t-1}+b_c); \\
c_t&=f_t \circ c_{t-1} + i_t \circ \tilde{c}_t; \\
h_t&=o_t \circ \sigma_h(c_t).
\end{aligned}
\end{equation}

where $x_t$ is the input vector; $f_t$ is the forget gate; $i_t$ is the input/update gate; $o_t$ is the output gate; $h_t$ is the output vector; $\tilde{c}_t$ is the cell input vector; $c_t$ is the hidden state vector; $W_f, W_i,W_o,W_c,U_f,U_i,U_o,U_c,b_f,b_i,b_o,b_c$ are trainable parameters.

$\sigma_{g}$, $\sigma_c$ and $\sigma_{h}$ are configurable activation functions, and we use $sigmoid$, $tanh$, $tanh$ by default.

where the operator $\circ$ denotes the Hadamard product (element-wise product). Operator $*$ denotes convolution for ConvLSTM\cite{xingjian2015convolutional}, and denotes matrix multiplication for LSTM\cite{hochreiter1997long}.

\subsubsection{GRU\cite{cho2014learning}}

\begin{equation}
\begin{aligned}
z_t&=\sigma_g(W_zx_t+U_zh_{t-1}+b_z); \\
r_t&=\sigma_g(W_rx_t+U_rh_{t-1}+b_r); \\
\tilde{h}_t &= \sigma_h(W_hx_t+U_h(r_t \circ h_{t-1}) + b_h); \\
h_t &= (1-z_t) \circ h_{t-1} + z_t \circ \tilde{h}_t.
\end{aligned}
\end{equation}

where $x_{t}$ is the input vector; $h_{t}$ is the output vector; $\hat{h}_{t}$ is the candidate activation vector; $z_{t}$ is the update gate vector; $r_{t}$ is the reset gate vector; $W_z,W_r,W_h,U_z,U_r,U_h,b_z,b_r,b_h$ are trainable parameters;
 $\sigma _{g}$, $\sigma_{h}$ are activations, and we use $sigmoid$ and $tanh$ by default.


\end{document}